\documentclass[10pt,twocolumn,letterpaper]{article}

\usepackage{cvpr}
\usepackage{times}
\usepackage{epsfig}
\usepackage{graphicx}
\usepackage{amsmath}
\usepackage{amssymb}
\usepackage{multirow}
\usepackage{subfigure}
\usepackage{interval}
\usepackage{dsfont}
\usepackage{chngpage}
\usepackage{xcolor,colortbl}

\usepackage[pagebackref=true,breaklinks=true,letterpaper=true,colorlinks,bookmarks=false]{hyperref}

\cvprfinalcopy % *** Uncomment this line for the final submission

\def\cvprPaperID{1152} % *** Enter the CVPR Paper ID here

\ifcvprfinal\fi

\newcommand*\rfrac[2]{{}^{#1}\!/_{#2}}
\newcommand{\layername}[1]{{\fontfamily{qcr}\selectfont#1}}
\newcommand{\myparagraph}[1]{{\vspace{0.5em} \noindent \bf #1}}
\begin{document}

%%%%%%%%% TITLE
\title{What Can Help Pedestrian Detection?}

\author{Jiayuan Mao\thanks{Equal contribution.}\hspace{0.4em}\thanks{Work was done during their internships at Megvii Inc.}\\
{\normalsize The Institute for Theoretical Computer Science (ITCS)}\\
{\normalsize Institute for Interdisciplinary Information Sciences}\\
{\normalsize Tsinghua University, Beijing, China}\\
{\tt\small mjy14@mails.tsinghua.edu.cn}
\and
Tete Xiao${}^*$${}^\dagger$\\
{\normalsize School of Electronics Engineering and Computer Science}\\
{\normalsize Peking University, Beijing, China}\\
{\tt\small jasonhsiao97@pku.edu.cn}
\and
Yuning Jiang\\
{\normalsize Megvii Inc.}\\
{\normalsize Beijing, China}\\
{\tt\small jyn@megvii.com}
\and
Zhimin Cao\\
{\normalsize Megvii Inc.}\\
{\normalsize Beijing, China}\\
{\tt\small czm@megvii.com}
}

% \author{Jiayuan Mao\thanks{Work was done during their internships at Megvii Inc.}\\
% {\normalsize Tsinghua University}\\
% {\normalsize Beijing, China}\\
% {\tt\footnotesize mjy14@mails.tsinghua.edu.cn}
% \and
% Tete Xiao${}^*$\\
% {\normalsize Peking University}\\
% {\normalsize Beijing, China}\\
% {\tt\footnotesize jasonhsiao97@pku.edu.cn}
% \and
% Yuning Jiang\\
% {\normalsize Megvii Inc.}\\
% {\normalsize Beijing, China}\\
% {\tt\footnotesize jyn@megvii.com}
% \and
% Zhimin Cao\\
% {\normalsize Megvii Inc.}\\
% {\normalsize Beijing, China}\\
% {\tt\footnotesize czm@megvii.com}
% }

\maketitle
\thispagestyle{empty}

\definecolor{Gray}{gray}{0.85}
\definecolor{White}{gray}{1}

% gray column and white column
\newcolumntype{a}{>{\columncolor{Gray}}c}
\newcolumntype{b}{>{\columncolor{white}}c}
\newcommand{\bfred}[1]{\color{red}{{#1}}}

\begin{abstract}
Aggregating extra features 
% (e.g. edges, segmentation, optical flows and disparity) 
has been considered as an effective approach to boost traditional pedestrian detection methods. 
However, there is still a lack of studies on whether and how CNN-based pedestrian detectors can benefit from these extra features. 
The first contribution of this paper is exploring this issue by aggregating extra features into CNN-based pedestrian detection framework. 
Through extensive experiments, we evaluate the effects of different kinds of extra features quantitatively. 
Moreover, we propose a novel network architecture, namely HyperLearner, to jointly learn pedestrian detection as well as the given extra feature. By multi-task training, HyperLearner is able to utilize the information of given features and improve detection performance without extra inputs in inference. The experimental results on multiple pedestrian benchmarks validate the effectiveness of the proposed HyperLearner.
\end{abstract}

\section{Introduction}
\label{sec:introduction}
Pedestrian detection, as the first and most fundamental step in many real-world tasks, \eg, human behavior analysis, 
gait recognition, intelligent video surveillance and automatic driving, has attracted massive attention in the last 
decade~\cite{dollar2009pedestrian,Geiger2012CVPR,dollar2009integral,zhang2015filtered,zhang2016faster,xiang2016subcategory}. 
However, while great progress has been made by deep convolutional neural 
networks (CNNs) on general object detection~\cite{ren2015faster,liu2015ssd,Dai2016aa,he2015deep}, 
research in the realm of pedestrian detection remains not as cumulative considering two major challenges.

\begin{figure}[!tb]
    \subfigure[]{
        \centering
        \includegraphics[width=1\linewidth]{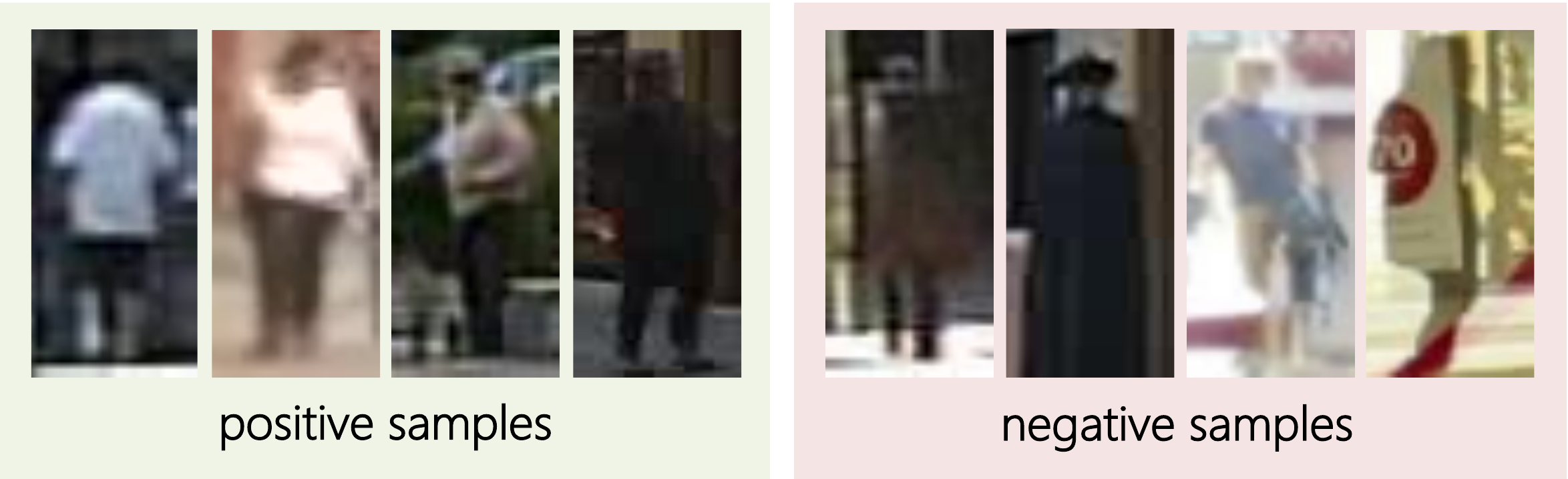}
        \label{fig:challengingcases1}
        \vspace{-1em}
    }
    \subfigure[]{
        \centering
        \includegraphics[width=1\linewidth]{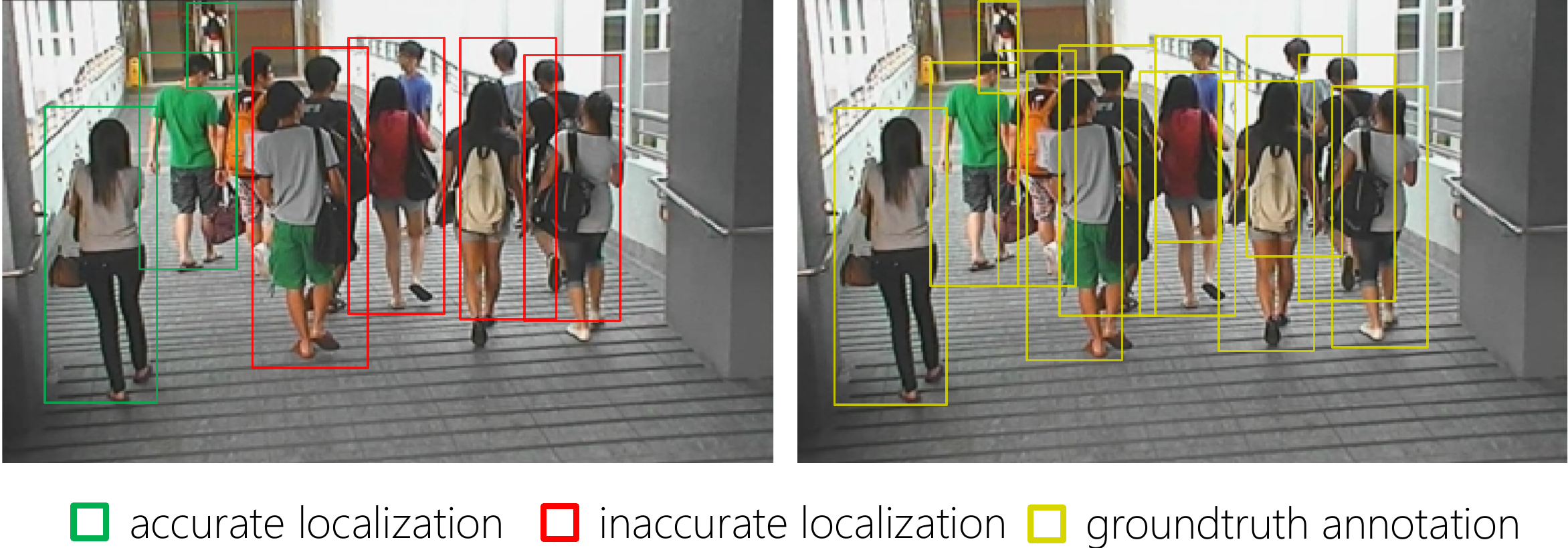}
        \label{fig:challengingcases2}
    }
    \caption[]{(a) Examples of true pedestrians and hard negative samples of low resolution. Without extra semantic contexts, 
    it is difficult to discriminate between them, even for human eyes.
    (b) Example of pedestrians in crowded scenes, where CNN-based detectors fail to locate each individual without 
    low-level apparent features.}
    \label{fig:challengingcases}
    \vspace{-1.8em}
\end{figure}

% this figure serves 3-channels.introduction
\begin{figure*}[!tp]
    \begin{center}
        \includegraphics[width=1.0\linewidth]{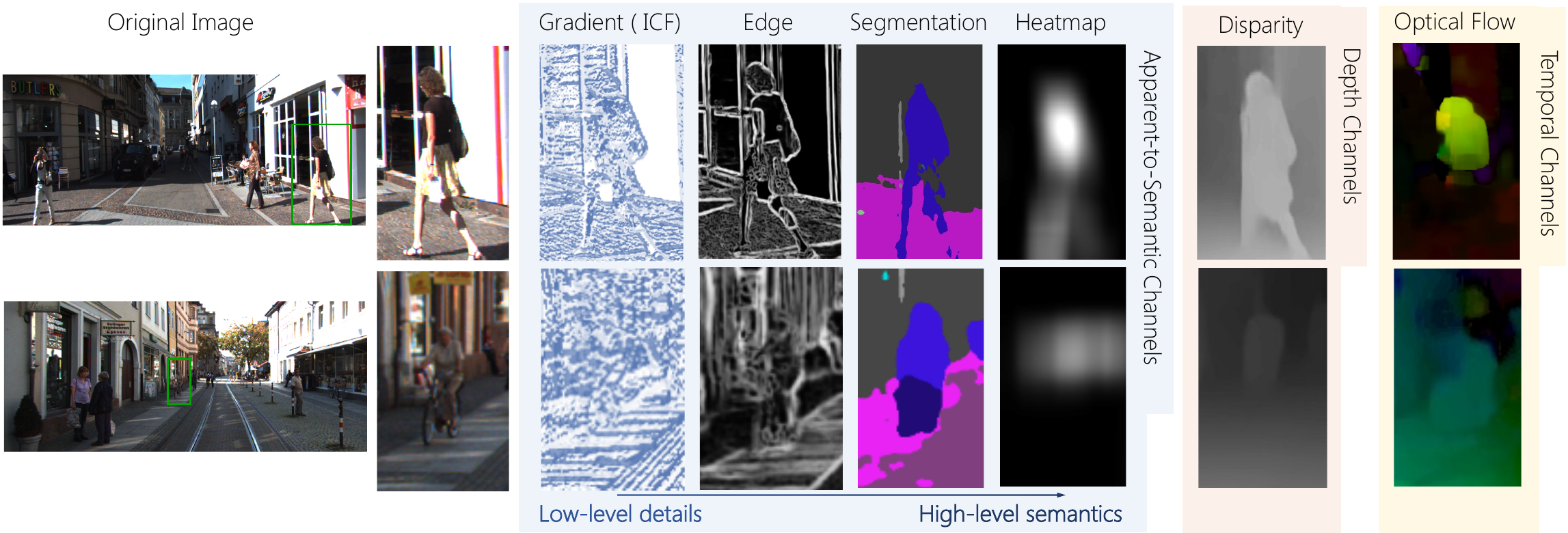}
        \vspace{-1em}
    \end{center}
    \caption{A demonstration of various channel features. Includes: apparent-to-semantic features, temporal features, depth features.}
    \label{fig:ChannelFeatureIntroduction}
    \vspace{-1em}
\end{figure*}

Firstly, compared to general objects, pedestrians are less discriminable from backgrounds. In other words, the discrimination relies more on 
the semantic contexts. As shown in Figure~\ref{fig:challengingcases1}, usually appearing in low resolution 
(less than $20{\times}40$ pixels), pedestrians together with the cluttered background bring about hard negative samples, such as 
traffic signs, pillar boxes, and models in shopping windows, which have very similar apparent features with pedestrians. 
Without extra semantic contexts, detectors working with such low-resolution inputs are unable to discriminate between them, 
resulting in the decrease in recall and increase in false alarms.

How to accurately locate each pedestrian is the second challenge.
Figure~\ref{fig:challengingcases2} is one showcase in practical applications where the pedestrians 
stand close in a crowded scene. 
As a result, detectors typically fail to locate each individual and hence produce a dozen of false positives due to 
inaccurate localization. 
This problem becomes even worse for CNN-based detectors since while convolution and pooling layers generate 
high-level semantic activation maps, they also blur the boundaries between closely-laid instances. 
An intuitive alternative to address the problem is to make use of extra low-level apparent features (\eg edges), 
for the purpose of solving the localization drawbacks by providing detectors with detailed apparent information.
% With this observation, To improve the localization accuracy, an intuitive approach which is somehow opposite to the earlier conclusion is proposed: extra low-level appearance features (\eg the edges and boundaries of pedestrians), rather than high-level semantic features, play a critical role in pedestrian detection.

In addition, in many applications, detectors can also benefit from other information, 
like depth when the camera is equipped with a depth sensor, or temporal information when a video sequence is input. 
However, it is still unclear how these information can be utilized by detectors, especially CNN-based detectors.

Given the observations above, one question comes up naturally: {\it what kind of extra features are effective and 
how they actually work to improve the CNN-based pedestrian detectors?} 
In this paper, we aim to answer this question and explore the characteristics of different extra features in 
pedestrian detection task. This paper contributes to:

\begin{adjustwidth}{1em}{0em}
    \noindent $\bullet$~Firstly, we integrate extra features as input channels into CNN-based detectors. 
    To investigate three groups of channel features: apparent-to-semantic channels, temporal channels and depth
    channels, extensive experiments are carried out on the KITTI pedestrian dataset~\cite{Geiger2012CVPR}. \par
    \noindent $\bullet$~Then, we experimentally analyze both advantages and disadvantages of different kinds of channel
    features.
    Specifically, we quantify the improvement brought by different channel features and provide insight into the error sources.\par
    % \noindent $\bullet$~Furthermore, to jointly learn pedestrian detection as well as a certain kind of channel feature,
    % a novel network architecture, namely HyperLearner, is proposed. 
    % In HyperLearner, channel features act as supervision instead of extra inputs. 
    % HyperLearner can learn the representations of channel features 
    % and utilize the information in them, while requiring no extra input in inference. 
    % We verify the effectiveness of HyperLearner on several pedestrian detection benchmarks and 
    % achieve state-of-the-art performance.
    \noindent $\bullet$~Moreover, a novel network architecture, namely HyperLearner, is proposed to aggregate extra
    features in a multi-task learning manner. In HyperLearner, channel features are aggregated as supervision instead of extra inputs,
    and hence it is able to utilize the information of given features and improve detection performance while requiring no
    extra inputs in inference.
    We verify the effectiveness of HyperLearner on several pedestrian detection benchmarks and 
    achieve state-of-the-art performance.
\end{adjustwidth}

\section{Related work}

Traditional pedestrian detectors, extended from Viola and Jones paradigm~\cite{viola2004robust}, 
such as \verb'ACF'~\cite{dollar2014fast}, \verb'LDCF'~\cite{nam2014local}, and 
\verb'Checkerboards'~\cite{zhang2015filtered}, filter various Integral Channels Features (ICF)~\cite{dollar2009integral} 
before feeding them into a boosted decision forest, predominating the field of pedestrian detection for years.
Coupled with the prevalence of deep convolutional neural network, CNN-based models~\cite{li2015scale,zhang2016faster,
cai2016unified} have pushed pedestrian detection results to an unprecedented level. 
% TODO:: i can't understand
In~\cite{zhang2016faster}, given region proposals generated by a Region Proposal Network (RPN),
CNN features extracted by an RoI pooling layer~\cite{girshick2015fast} are fed into a boosted forest; 
while in Cai \etal ~\cite{cai2016unified}, a downstream neural network architecture is proposed to 
preform end-to-end detection.

Integrating channel features of different types has been proved to be useful in many decision-forest-based pedestrian 
detectors. Prior work by Park \etal ~\cite{park2013exploring} embeds optical flow into a boosted decision forest to 
improve pedestrian detectors working on video clips. \verb'CCF'~\cite{yang2015convolutional} 
uses the activation maps of a VGG-16~\cite{simonyan2014very} network 
pretrained on ImageNet~\cite{krizhevsky2012imagenet} as channel feature, while Costea and 
Nedevschi~\cite{daniel2016semantic} utilize the heatmap of semantic scene parsing, in which detectors benefit from the 
semantic information within a large receptive field. 
% there has been a lack of attention on this problem regarding 
However, the problem whether and how CNN-based pedestrian detectors can benefit from extra features still exhibits 
a lack of study.

\section{Channel features for pedestrian detection}
In this section, we empirically explore the performance boost when extra channel features are integrated into CNN-based detectors.
% We first describe the benchmark, evaluation metrics and baseline detector we use.
% After introducing the selected channels in Section~\ref{subsec:introtochannel}, we experiment explore the different network designs for integrating channel features in Section~\ref{subsec:integration}.
% Finally, in Section~\ref{subsec:comparison}, a detailed analysis on the effects of selected channel features is provided.
% In this section we explore the performance boost when extra channel features are integrated into CNN-based detectors.
% the information in which vary from local to global and detailed to semantic. 
% We also provide meticulous analysis on their effects on CNN-based detectors.

\subsection{Preliminaries} 
\label{subsec:channelfeaturepreliminaries}
Before delving into our experiments, we first describe the dataset, evaluation metrics and baseline detector we use. 

\myparagraph{KITTI dataset}
We choose KITTI dataset~\cite{Geiger2012CVPR} for channel feature analysis considering its possession of 
pedestrians of various scales in numerous scenes, as well as the information of adjacent frames and stereo data.
KITTI contains $7,481$ labeled images of resolution $1250{\times}375$ and another $7,518$ images for testing. 
The training set is further split into two independent set for training and validation following~\cite{chen20153d}.
% It is guaranteed that samples from the same video clip will not be assigned to both set. 
% ``Person'' class evaluation in KITTI is divided into two classes: pedestrian and cyclist with three levels: easy, moderate and hard, defined by minimum bounding-box height, maximum occlusion level and maximum truncated ratio. 
The person class in KITTI is divided into two sub-classes: pedestrian and cyclist, both evaluated under PASCAL criteria~\cite{everingham2010pascal}. 
% For both pedestrians and cyclists an overlap of $50\%$ is required. 
KITTI contains three evaluation metrics: {\it easy, moderate} and {\it hard}, 
with difference in the \emph{min$.$} bounding box height, \emph{max$.$} occlusion level, \etc.
% with three metrics: easy, moderate and hard~\cite{?}.
Standard evaluation follows moderate metric.
% We use moderate metric following official standard.

\myparagraph{Faster R-CNN} 
Our baseline detector is an implementation of Faster R-CNN~\cite{ren2015faster},
initialized with VGG-16~\cite{simonyan2014very} weights pretrained on ImageNet~\cite{krizhevsky2012imagenet}.
It consists of two components: a fully convolutional Region Proposal Network (RPN) for proposal generation, 
and a downstream Fast R-CNN (FRCNN) detector taking regions with high foreground likelihood as input. 

Since KITTI contains abounding small objects, we slightly modify the framework as~\cite{xiang2016subcategory} and~\cite{cai2016unified}.
Specifically, we adjust the number of anchors from $3$ scales and $3$ ratios to $5$ scales and $7$ ratios; 
besides, all \layername{conv5} layers are removed to preserve an activation map of high resolution for both RPN and FRCNN.
% There is no extra modification.

We choose Faster R-CNN not only for its prevalence and state-of-the-art performance, 
but also generality: our observations should remain mostly effective when similar techniques are applied 
in other CNN-based pedestrian detectors.

\subsection{Introduction to channel features}
\label{subsec:introtochannel}
In this section, we introduce the channel features we integrated into the CNN-based pedestrian detector.
Based on the type of information they carry, the selected channel features for integration are divided into 
three groups: apparent-to-semantic channels, temporal channels and depth channels. 
Figure~\ref{fig:ChannelFeatureIntroduction} provides a demonstration of all channels.

\begin{figure}[!tb]
\begin{center}
   \includegraphics[width=0.9\linewidth]{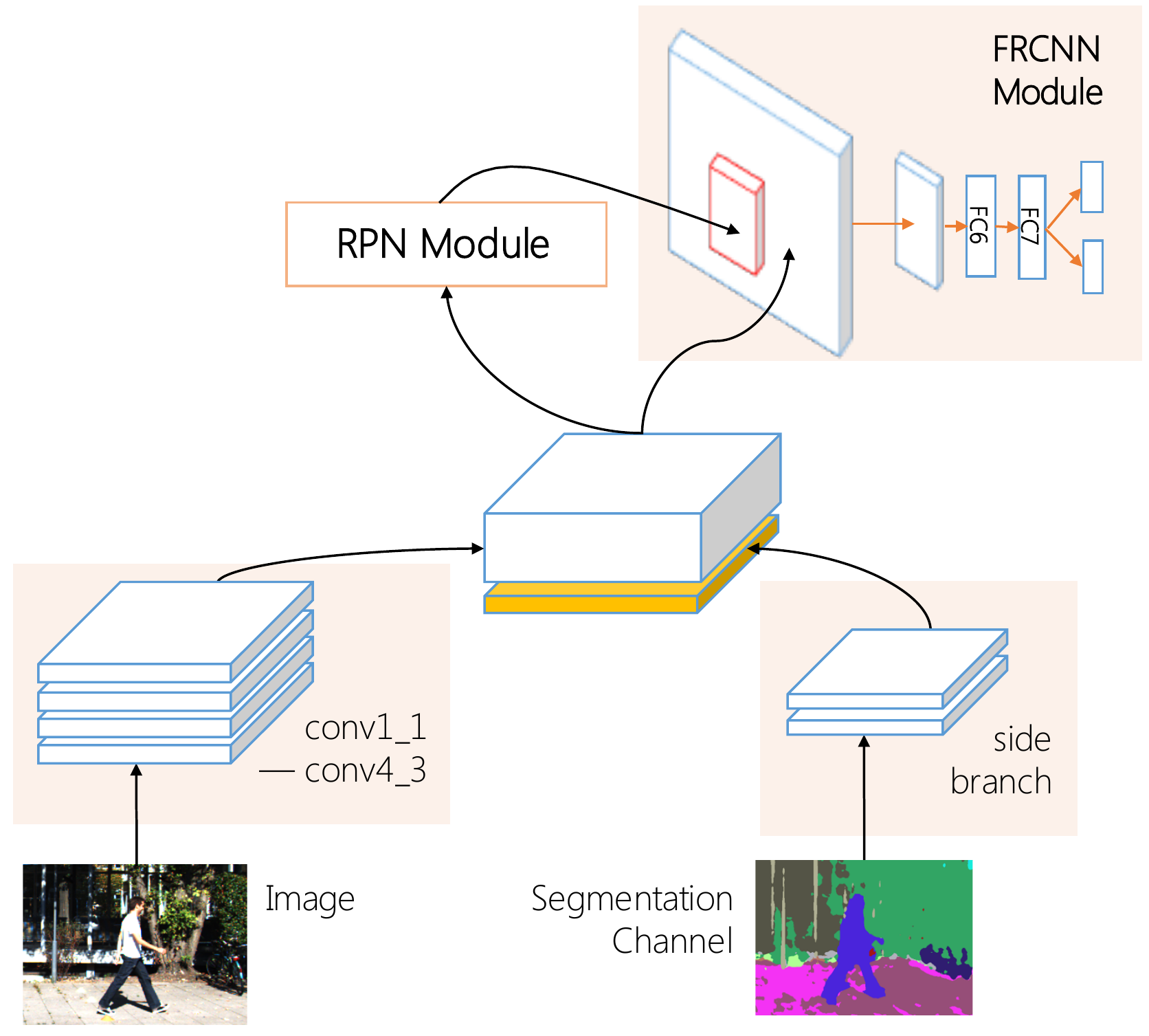}
\end{center}
   \caption{As described in Section~\ref{subsec:integration}, our Faster R-CNN for channel feature integration. 
    The side branch takes channel features as input and generates channel feature representations before concatenated
    with \layername{conv4\_3}.}
\label{fig:FasterPipeline}
\vspace{-1em}
\end{figure}

\myparagraph{Apparent-to-semantic channels} 
This group of channels includes ICF channel~\cite{dollar2009integral}, edge channel, segmentation channel and heatmap channel. 
The information in these channels ranges from low-level apparent to high-level semantic.

The ICF channel is a handy-crafted feature channel composed of 
LUV color channels, gradient magnitude channel, and histogram of gradient (HOG) channels, 
which has been widely employed in the decision-forest-based detectors~\cite{dollar2014fast,nam2014local,zhang2016far}.
Containing only colors and gradients within a local patch, ICF channel represents the most low-level but 
detailed information of an image.

The edge channel is extracted from the second and third layers of HED network~\cite{xie15hed}.
% a contemporary edge detector which performs image-to-image predictions for edges. 
Different with traditional edge detector such as Canny~\cite{canny1986computational}, 
the HED framework produces more semantically meaningful edge maps (see Figure~\ref{fig:ChannelFeatureIntroduction}). 
The edge channel is thus considered as a mid-level feature channel containing both detailed appearance as well as high-level semantics.

As in~\cite{long2015fully,chen2014semantic}, a fully convolutional network (FCN) is trained on 
MS-COCO dataset~\cite{lin2014microsoft} to generate the semantic segmentation channel, 
where each pixel represents the probability of the category (\eg, person and street) it belongs to. 
The segmentation channel carries higher-level semantic information, while still perserving some detailed appearance
features, \ie, the boundaries between objects of different categories.
% Although most low-level details are lost in the segmentation channel, 
However, two closely-laid instances of same category cannot be distinguished from each other in the segmentation 
channel without contour of each instance.

Furthermore, to obtain a feature channel with only high-level semantics, 
we blur the segmentation channel into the heatmap channel. 
By doing so, the clear boundaries between objects of different categories are also removed and 
only high-level information of categories remains. 
% In the following experiments we will compare the blurred heatmap channel with the segmentation channel to verify the importance of the low-level appearance features in pedestrian detection.

\myparagraph{Temporal channels} 
The temporal features (\eg, optical flow~\cite{beauchemin1995computation} and motion~\cite{wang2009evaluation}) 
have been proved to be beneficial to traditional pedestrian detectors~\cite{walk2010new,park2013exploring} working on videos. 
% Now we are wondering whether the temporal features can help for the CNN-based framework. 
To test their effectiveness in CNN-based framework, we extract optical flow channel as representative using temporally adjacent frames.

\myparagraph{Depth channels} 
With more and more depth sensors employed in intelligent systems such as robotics and automatic driving, 
the depth information available in these tasks becomes an alternative extra channel feature to boost detectors. 
% Here we also explore leveraging the depth channels in the pedestrian detection problem. 
Instead of using the sparse point clouds captured by laser radars, we turn to DispNet~\cite{mayer2015large} 
to reconstruct the disparity channel from stereo images.

\subsection{Integration techniques}
\label{subsec:integration}
% Before analysis the performance of each channel feature, we first discuss the network design for channel feature integration.
% To incorporate additional channel features into our baseline Faster R-CNN framework is a complicated problem, because there are plenty of ways to inject extra features in any stage.
We integrate channel features by creating a new shallow side branch alongside the VGG-16 main stream 
(see Figure~\ref{fig:FasterPipeline}). 
% To make fully use of the pretrained features (\ie, the VGG-16 weights), we integrate channel features by creating a new shallow side branch besides the deep feature extraction layers 
This side branch consists of several convolution layers (with kernel size $3$, padding $1$ and stride $1$) and max
pooling layers (with kernel size $2$ and stride $2$), outputing an $128$-channel activation maps of $\rfrac{1}{8}$ input size, 
which is further concatenated with activation map \layername{conv4\_3}. 
The concatenated activation map is fed into the RPN and FRCNN to preform detection.
% Each convolution layer in the side branch has $3{\times}3$ kernel size with stride $1$ and $1$ pixel padding, 
% followed by a $2{\times}2$ max-pooling layer with stride $2$. 
% The input channel feature map is resized in advance using bilinear interpolation to ensure that activation maps being concatenated possess the same size. 

We experiment different compositions of the side branch: the number of convolution layers and the initial weights
(\ie, a random gaussian kernel, or pretrained weights). 
The technique we employed to pretrain the side branch is to train a Faster R-CNN detector which completely relies on 
the side branch and intialize the side branch with the weights from this network. 

\begin{table}[!htbp]
\begin{center}
\setlength{\tabcolsep}{5pt}
\begin{tabular}{c|c|c||a c c}
\multicolumn{3}{c||}{Model} & \multicolumn{3}{c}{Pedestrian} \\ \cline{1-6}
    & \#Convs & Init$.$~W$.$ & \multicolumn{1}{c}{Mod} & Easy & Hard\\ 
\hline
\hline
O & N/A & N/A & 68.96 & 73.33 & 60.43 \\
\hline
A & 2 & random & \bfred{70.80} & \bfred{78.15} & \bfred{62.16} \\
B & 1 & random & 70.40  & 75.17 & 61.92 \\
C & 2 & pretrained & 69.92 & 77.33 & 61.65 
\end{tabular}
\end{center}
\caption{Detection improvement by integrating channel features on KITTI validation set. Model ``O'' is our baseline
detector. ``\#Convs" means the number of convolution layers in the side branch. ``Init$.$~W$.$''
denotes initial weights for the side branch. The input images are not enlarged.}
\label{tab:segfeatureadding}
\end{table}

Summariesed in Table~\ref{tab:segfeatureadding}, all integration methods improve the baseline Faster R-CNN detector 
in KITTI validation set on both classes across all three metrics. 
Nevertheless, the model with two extra convolution layers outperforms the model with only one extra convolution layer.
% which is consistent with the fact that a deeper network possesses more ability of encoding information.
A pretrained side branch does not perform well when further assembled with the VGG-16 network. 
When probing the network, we find that the model with pretrained weights tend to ``rely'' more on the sidebranch,
(\ie, activation map produced by side branch has much greater value than the main stream).
Given the fact that the side branch was pretrained to perform detection independently, this inbalance may be a 
cause accounting for the performance degradation. 
% Given the fact that the side branch was pretrained to perform detection independently, 
% the model will ``rely'' more on the side branch 
Based on the analysis, we use two convolution layers with random Gaussian initialization in all future experiments.

\begin{table}[!t]
\begin{center}
\setlength{\tabcolsep}{4.5pt}
\begin{tabular}{l || c  c  c  c}
\multirow{2}{*}{Model} & \multicolumn{4}{c}{Recall} \\ \cline{2-5}
& $\left(0, 80\right]$ & $\left(80,160\right]$ & $\left(160,\inf\right]$ & all scales\\ 
\hline
\hline
Baseline       & 21.3\% & 87.6\% & 96.8\% & 70.0\% \\
\hline
+Segmentation   & 35.6\% & 88.2\% & 96.8\% & 74.0\% 
\end{tabular}
\end{center}
\caption{Recall comparison at 70\% precision between baseline and segmentation channel at different pedestrian heights. The results are based on 1x scale. }
\label{tab:recallcomparison1x}
\vspace{-1em}
\end{table}

\subsection{Comparison and analysis}
\label{subsec:comparison}
% To analyze the effectiveness of integrating various channel features, 
We conduct experiments on two input scales 
(\emph{1x} and \emph{2x}). 
Table~\ref{tab:featurecomparison} summarizes the results.
For a fair comparison, a controlled experiment in which the original image is used as input of the side branch is 
also included. 
% : ICF, edge, segmentation and heatmap channels (apparent-to-semantic); disparity channel (depth); optical flow channel (temporal). 
% We also list their average improvement across all three metrics.

% TODO:: why ICF doesnot work
In general, models integrated with extra channel features show improvement over the baseline. 
The experiment using original image as extra input shows nonobvious improvement, 
which confirms that the performance gain is indeed attributed to channel feature integration. 
Among all channel features, ICF channel shows least contribution to the detection performance in both scales. 
We conjecture the reason is that in deep convolutional networks, CNN features are more discriminative than 
hand-crafted features like HOG.

Recall the two major challenges for pedestrian detection: hard negative samples and the individual localization.
Through detailed analysis, we demonstrate how CNN-based detectors can benefit from extra channel features 
to overcome these problems.

\myparagraph{{\emph 1x} experiments} In \emph{1x} experiments, 
% When incorporated into the baseline detector, heatmap channel (\ie blurred segmentation) brings similar gains as segmentation channel with object boundaries. 
channels that carry more semantic information show better performance.
As shown in Table~\ref{tab:featurecomparison}, detectors with segmentation channel and heatmap channel 
% outperform other channel features, and 
bring most significant improvement to the detector. 
% both show significant improvement compared to the baseline detector. 
In accord with our previous hypotheses, the detectors utilize the semantic context provided by extra channel features 
to discriminate pedestrian of low resolution from hard negative samples.
% This is in accord with our previous hypotheses that semantic context is helpful for CNN-based detectors with inputs of low resolution.
%This hierarchical result is accord with apparent-to-semantic channel features, where a channel with more semantic information and less details tend to boost baseline detector for larger margins. 
% High-level channel features, which contain more semantic information, can significantly improve the recall. 

Table~\ref{tab:recallcomparison1x} provides the recall comparison at certain precision rate ($70\%$) between models 
with segmentation channel and the baseline model for pedestrians of different sizes.
All pedestrians are divided into four groups based on their heights in pixel. 
Leading absolute $4\%$ recall rate on average, the detector with segmentation channel performs significantly better 
in recall for small pedestrians (less than or equal to $80$ pixel in height).
% while showing slight advantage over the baseline in terms for large pedestrians. 
% shows that segmentation channel mainly improves performance on small objects.

% \begin{figure}[!tb]
% \begin{center}
%    %\includegraphics[width=0.8\linewidth]{egfigure.eps}
% \end{center}
%    \caption{Precision-Recall curve for baseline and segmentation at 1x scale}
% \label{fig:prcurve1x}
% \end{figure}

\begin{table*}[!hbt]
\begin{center}
\setlength{\tabcolsep}{3.8pt}
\begin{tabular}{l || a c c | a c c | a || a c c | a c c | a}
\multirow{2}{*}{Model} & \multicolumn{3}{c|}{Pedestrian 1x Input} & \multicolumn{4}{c||}{Improvement} 
& \multicolumn{3}{c|}{Pedestrian 2x Input} & \multicolumn{4}{c}{Improvement} 
\\ \cline{2-15}
    & \multicolumn{1}{c}{Mod} & Easy & Hard & \multicolumn{1}{c}{Mod} & Easy & Hard & \multicolumn{1}{c||}{Avg}
    & \multicolumn{1}{c}{Mod} & Easy & Hard & \multicolumn{1}{c}{Mod} & Easy & Hard & \multicolumn{1}{c}{Avg}\\
\hline
\hline
Fr-RCNN*~\cite{ren2015faster}    & 59.29 & 64.53 & 53.01 & - & - & - & -
                                                 & 71.05 & 76.00 & 62.08 & - & - & - & - \\
MS-CNN~\cite{cai2016unified}                & 68.37 & 73.70 & 60.72 & - & - & - & -
                                                & 72.26 & 76.38 & 64.08 & - & - & - & - \\
\hline
Our Baseline    & 68.96 & 73.33 & 60.43 & - & - & - & -
                    & 71.21 & 77.73 & 62.19 & - & - & - & -\\
+ Original img   & 68.63 & 76.61 & 60.45 & -0.33 & +3.28 & +0.02 & +0.99
                   & 71.33 & 76.72 & 62.17  & +0.12 & -1.01 & -0.02  &  -0.30 \\
+ ICF              & 68.40  & 73.56 & 60.20 & -0.56 & +0.23 & -0.23 & -0.19
                  & 71.80   & 77.40 & 62.79 & +0.59 & -0.33 & +0.60 & +0.29 \\
+ Edge        & 69.49   & 76.28 & 60.89 & +0.53 & +2.95 & +0.46 & +1.31
                  & 72.34   & 78.32 & 63.28 & +1.13 & +0.59 & +1.09 & +0.94 \\
+ Segmentation & \bfred{70.80} & \bfred{78.15} & \bfred{62.16} & \bfred{+1.84} & \bfred{+4.82} & \bfred{+1.73} & \bfred{+2.80}
                    & \bfred{72.54} & \bfred{78.49} & \bfred{63.61} & \bfred{+1.33} & \bfred{+0.76} & \bfred{+1.42} & \bfred{+1.17} \\
%+ Edge \& Segmentation & 70.47 & 76.12 & 62.15 & 52.59 & 76.54  & 50.73
%                           & 71.66 & 76.97 & 63.08 & \bfred{67.92} & 83.55 & 63.18 \\
%+ Heatmap      & 70.94 & \bfred{78.89}    & 62.10 & 52.6 & 76.90    & \bfred{50.25}
%                   & 70.19 & 74.06 & 61.53 & 65.69  & \bfred{85.02} & 62.92 \\
+ Heatmap   & 70.33 & 78.03 & 61.75 & +1.37 & +4.70 & +1.32 & +2.46
                & 71.39 & 77.64 & 62.34 & +0.18 & -0.09 & +0.15 & +0.08 \\
\hline
+ Disparity         & 70.03 & 77.74 & 61.48 & +1.07 & +4.41 & +1.05 & +2.18
                    & 71.72 & 77.52 & 62.47 & +0.51 & -0.21 & +0.28 & +0.19 \\
+ Optical Flow   & 69.39 & 77.07 & 60.79 & +0.43 & +3.74 & +0.36 & +1.51
                    & 71.13 & 76.85 & 62.24 & -0.08 & -0.88 & +0.05 & -0.25 \\
\end{tabular}
\end{center}
\caption{Channel features comparison on KITTI validation set.
We list improvement across all three KTTTI metrics as well as the average. 
    *: Our reproduced Faster R-CNN with same parametrs as in \cite{ren2015faster}. 
    The baseline is a re-implementation of Faster RCNN pipeline, consisting of slight
    differences with the basic Faster RCNN (See Section~\ref{subsec:channelfeaturepreliminaries}).
}
\label{tab:featurecomparison}
\vspace{-1em}
\end{table*}

\begin{figure}[!tb]
    \begin{center}
    \subfigure[False positive sources at $70\%$ recall rate]{
        \centering
        \includegraphics[width=0.85\linewidth]{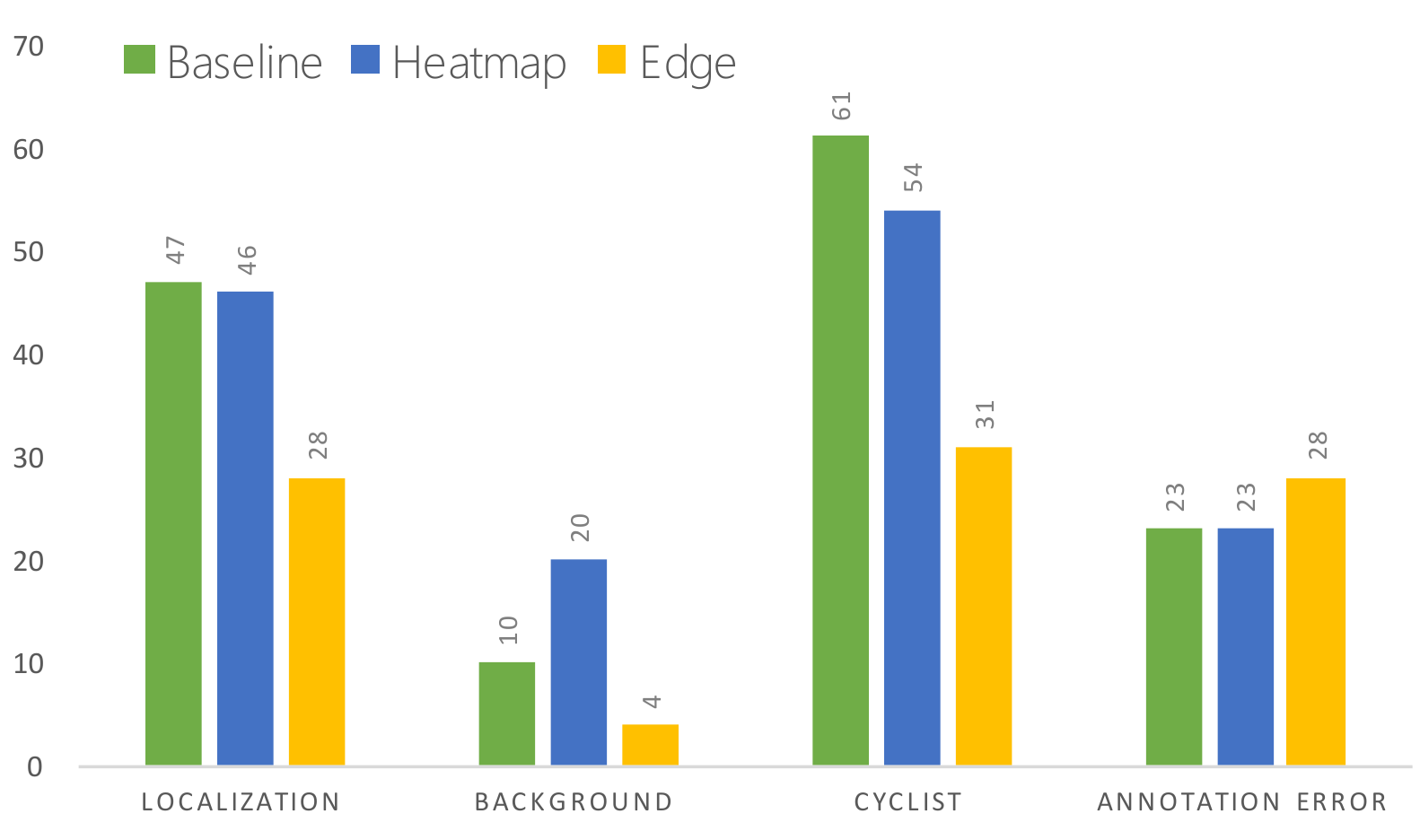}
        \label{fig:fpanalysis1}
    }
    \subfigure[Top-200 false positives sources]{
        \centering
        \includegraphics[width=0.85\linewidth]{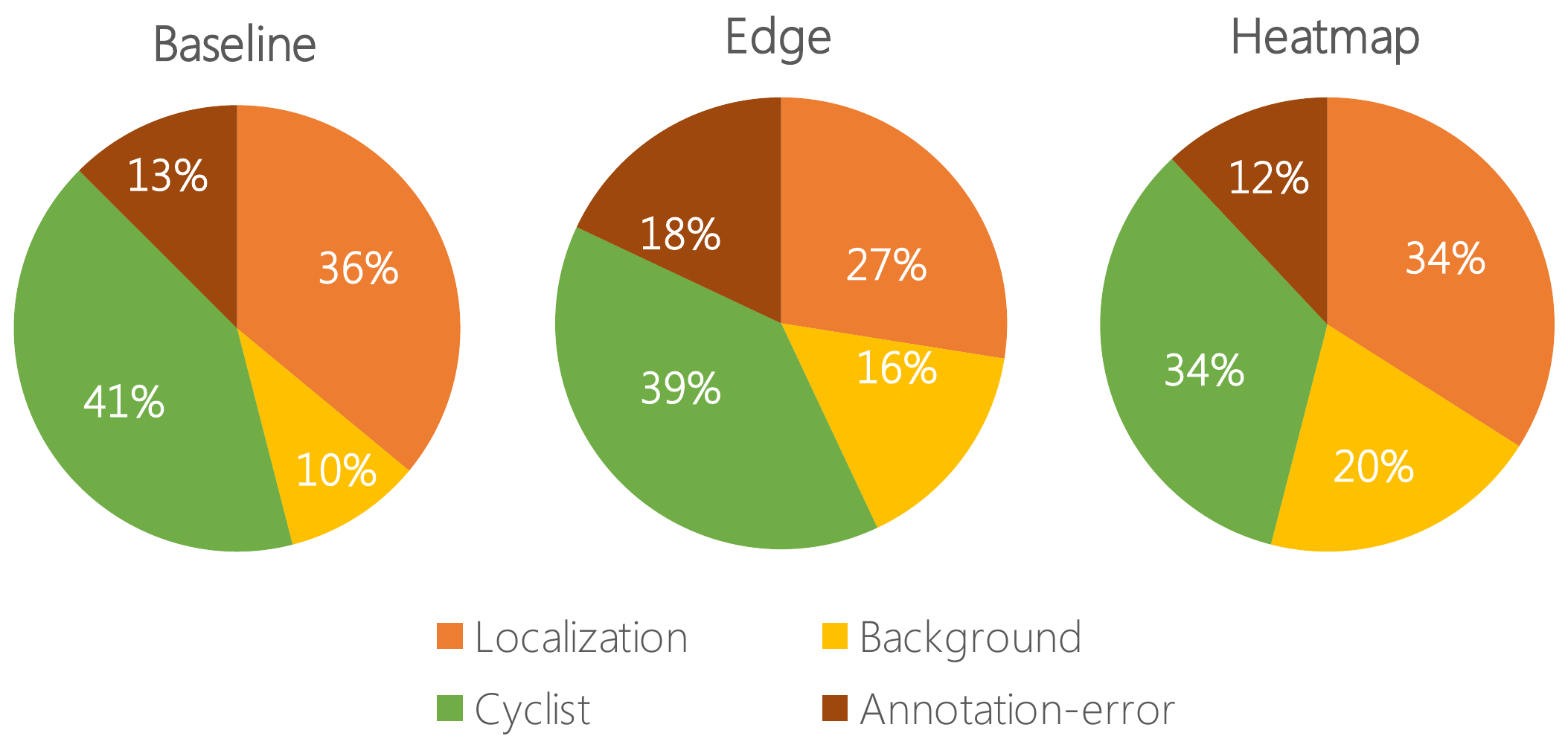}
        \label{fig:fpanalysis2}
    }
    \end{center}
    \caption{False positive analysis for baseline, edge channel and heatmap channel at \emph{2x} scale.
All false positives are categorized into four types: localization error, background classification error, 
cyclist classification error, and annotation error. 
Localization error is defined as non-matched detection bounding boxes which overlap with a groundtruth but iou~$<0.5$,
% which overlap with one of the groundtruth bounding boxes, 
while background error has no overlap with any groundtruth box. 
% In KITTI dataset, cyclist class is considered different from pedestrian class. 
% Therefore we label bounding boxes with cyclist error when they match cyclist groundtruth bounding boxes. 
Cyclist error happens when a bounding box match cyclist groundtruth.
Annotation error occurs when detection ``matches'' a {\it de facto} groundtruth which, however, is not annotated.}
    \vspace{-1em}
\end{figure}

\myparagraph{{\emph 2x} experiments} 
In \emph{2x} experiments, model with only high-level semantic information but no low-level apparent features 
(\ie the heatmap channel) fails to produce consistent improvement over the baseline model compared to the \emph{1x} experiments. 
Nonetheless, channel features with both high-level semantic and low-level apparent information 
(edge channel and segmentation channel) outperforms other channels.
A possible explanation for this is that when it comes to large input scale, low-level details 
(\eg, edge) will show greater importance in detection. 
To further explore this phenomenon, we randomly sampled $\rfrac{1}{4}$ of images (about $800$) 
from validation set and collected false positive statistics at $70\%$ recall rate, as shown in Figure~\ref{fig:fpanalysis1}. 
% Cyclist classification errors form the most common error source in all three models. 
% Localization error, which usually occurs with double detections on the same pedestrian, is also a serious problem.
While in Figure~\ref{fig:fpanalysis2}, we also count top-200 false positives in the validation set and show the 
fractions of each error source. 
Not only inhibiting false positives across all categories at a high recall, edge channel also contributes 
significantly to the localization precision. 
Integrated with the edge channel, detector lowers localization error rate by absolute $9\%$ and $7\%$ compared with 
the baseline and the detector with heatmap channel respectively. 
This proves that 
% besides helping detectors to discriminate pedestrians from backgrounds,
% and other categories with 
% similar appearance features (\eg, cyclists, traffic signs and pillar boxes), 
channel features with low-level apparent features (\eg, boundaries between individuals and contours of objects) 
improve localization precision when the input image is of high resolution. 

% It is also worth noticing that ICF channel does not produce any improvement in both scales. This may due to the fact that which provides evidence that, 
% once-dominated \emph{pure} low-level features, such as LUV color channels and HoG channels, are learned automatically by neural network. This interesting finding provides more insights toward hand-crafted features and gaps between traditional pedestrian detectors and neural network based detectors.

\begin{figure*}[!tb]
    \centering
    \includegraphics[width=0.9\linewidth]{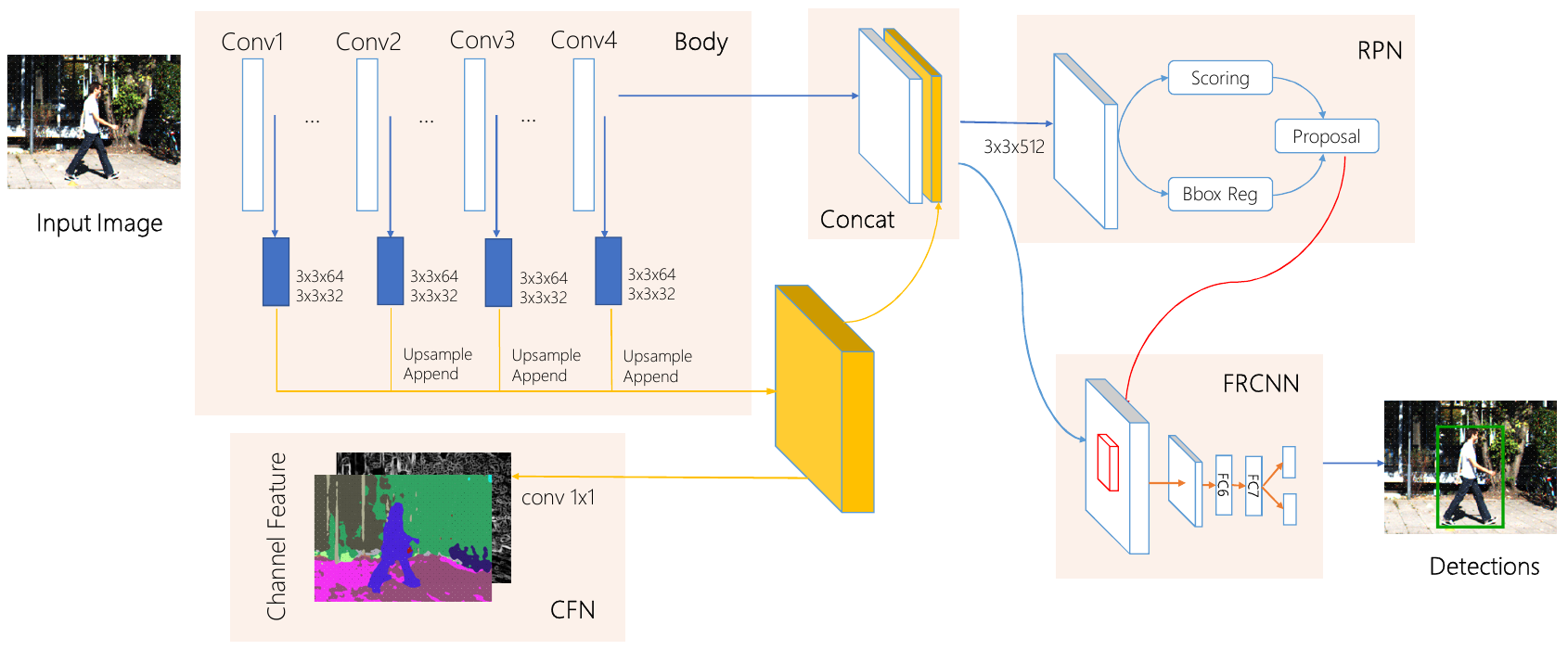}
    \caption{The proposed HyperLearner, which consists of 4 components: body network, channel feature network (CFN), 
    region proposal network (RPN) and Fast R-CNN (FRCNN). HyperLearner learns representations of channel features while 
    requiring no extra input in inference. Refer to Section~\ref{subsec:hyperlearner} for details.}
    \label{fig:hyperlearner}
\end{figure*}

Besides, We witness noticeable improvement in \emph{1x} when optical flow is integrated into the detector. 
Park \etal~\cite{park2013exploring} also proved this effectiveness in decision-forest-based detectors with a detailed
analysis. 
% The slight performance drop in 2x scale may be ascribed to the unstability as analyzed in~\cite{park2013exploring}.
For the disparity channel, the results are very similar to the results of heatmap channel. To have an insight into this, 
we should notice that the relative value in a disparity map also serves as a ``segmentation-like'' channel (see
Figure~\ref{fig:ChannelFeatureIntroduction}), while the 
absolute value has only limited effects compared to the deep convolutional features and the predefined anchors. 
%which are strong priors about the size of the objects in the image.

% The above experiments have shown that channel features can boost a detector working on images of both low resolution 
% and high resolution.
% With these channel features, we can narrow most of the gap between resolutions without introducing heavy computational 
% cost brought by enlarging the input image, and push state-of-the-art forward. 

\newcommand{\Loss}{\mathcal{L}}

\section{Jointly learn the channel features}
As observed above, integrating channel features into the network can boost our detector working on images of both low 
resolution and high resolution.
With these channel features, we can narrow most of the gap between resolutions without introducing heavy computational 
cost brought by enlarging the input image, and push state-of-the-art forward. 

However, a brute-force integration method is computationally expensive with respect to the basic Faster R-CNN, 
given that the input channel feature usually requires extra computational cost.
While many of the channel features comes from neural networks (\eg, semantic segmentation and edge), 
it is natural to think of ``teaching'' our neural-network both channel features generation and detection.
In the following section, we propose a new network structure to address the issue in a multi-task learning manner, 
namely, HyperLearner.
% , and present our training methods as well as experiments.

\subsection{HyperLearner}
\label{subsec:hyperlearner}
The HyperLearner framework is illustrated in Figure~\ref{fig:hyperlearner}. As shown, our system consists of four components: 
the body network for activation map generation, a channel feature network (CFN), a region proposal network (RPN) and 
a Fast R-CNN (FRCNN) network for final detection task.

From the very left, the entire image is forwarded through multiple convolution layers to generate the hierarchical 
activation maps. We first aggregate activation maps and make them into a uniform space, namely aggregated activation map. 
Aggregating activation maps from multiple level has been proved to be useful and important in many computer vision 
tasks~\cite{kong2016hypernet,xie15hed} for its ability to collect rich hierarchical representations.
This aggregated map is then fed into the channel feature network (CFN). CFN is a feed-forward fully convolutional 
network (FCN) for channel feature prediction. Unlike Faster R-CNN, RPN and FRCNN do not only take the output of the last 
convolution layer (\layername{conv4\_3}) as input. 
Instead, the aggregated activation map is also fed into the RPN, as well as FRCNN. By sharing the same aggregated 
activation map, the RPN and FRCNN are able to benefit from the representations CFN learned.

\myparagraph{Aggregated activation map}
The body network takes the raw image, of shape $3{\times}H{\times}W$, as its input, and outputs several activation maps. 
In our experiments, the body network is a VGG-16~\cite{simonyan2014very} network (without \layername{conv5\_1} to 
\layername{conv5\_3}) intialized with the weights pretrained on ImageNet~\cite{krizhevsky2012imagenet}. 
We extract the activation maps from layer \layername{conv1\_2}, \layername{conv2\_2}, \layername{conv3\_3} and 
\layername{conv4\_3}. Due to the pooling layer in the network, these maps are of different size and number of channels. 
We add two convolution layers after each map and keep their numbers of output channels same ($32$ in all our experiments). 
The high-level maps are then upsampled to the same size as the first activation map. 
Finally, they are concatenated together to form the aggregated activation map.

\myparagraph{Channel Feature Network (CFN)}
The CFN directly takes the aggregated activation map to generate the predicted channel feature map through 
a fully convolutional structure. This map is typically of the same shape as the raw image. 
For example, the predicted channel feature may be a semantic segmentation map of several categories, 
or an edge detection map like HED Network~\cite{xie15hed}.

\myparagraph{Region Proposal Network (RPN) and Fast-RCNN (FRCNN)}
We build the RPN and FRCNN using the same structure as proposed in~\cite{ren2015faster}. RPN and FRCNN now take both 
last convolutional activation map in the VGG16 network (\layername{conv4\_3}) and the aggregated activation map from 
the body network as the inputs. The proposals generated by RPN are then fed into FRCNN to perform final detection.

\subsection{Training Details}
\myparagraph{Loss Function}
During the training phase, besides the raw image and groundtruth bounding boxes for standard Faster R-CNN framework, 
the HyperLearner also takes a channel feature map as its supervisor, which is typically generated by another CNN 
(\eg, semantic segmentation and edge). To address the channel feature learning, we introduce a new pixel-level loss. 
Denote the feature map predicted by the CFN as $C_{x, y}$, and the supervisor map as $S_{x, y}$. 
The loss is computed by:
$ \displaystyle \frac{1}{H \times W} \sum_{(x, y)} \ell(S_{x, y}, C_{x, y}), $
where $H$ and $W$ represents the size of the feature map and $\ell$ is a loss function for a single pixel. 
In binary probabilistic maps, like edge map, cross-entropy loss is used, given by:
$ \ell(p, q) = \beta_{x, y} {\big(} -p \log q - (1-p) \log (1-q) {\big)}, $
where $\beta$ is a weight function to balance the positive labels and negative labels. 
If $S_{x, y} > 0.5$, $\beta = 1 - |S_{+}|\Large/|S|$; otherwise, $\beta = |S_{+}|\Large/|S|$, where $|S_{+}| = \sum \mathds 1[S_{x, y} > 0.5]$.
For multi-class probabilistic maps, like segmentation map, cross-entropy loss is used. 
For other tasks, MSE loss is used.

The final loss for the network is thus computed by:
% \begin{eqnarray*}
% \Loss = \Loss_{\mathrm{CFN}} + \lambda_1 \Loss_{\mathrm{RPNcls}} + \lambda_2 \Loss_{\mathrm{RPNbbox}} & & \\
% + \lambda_3 \Loss_{\mathrm{FRCNNcls}} + \lambda_4 \Loss_{\mathrm{FRCNNbbox}} & &
% \end{eqnarray*}
$ \Loss = \Loss_{\mathrm{CFN}} + \lambda_1 \Loss_{\mathrm{RPNcls}} + \lambda_2 \Loss_{\mathrm{RPNbbox}}
+ \lambda_3 \Loss_{\mathrm{FRCNNcls}} + \lambda_4 \Loss_{\mathrm{FRCNNbbox}}$
where the last four component remains the same as Faster R-CNN~\cite{ren2015faster}. In all our experiments, we set all $\lambda_i = 1$.

\myparagraph{Multi-stage training}
The aggregated activation map acts as an important role in the framework, which must be carefully trained. 
% Instead of trickily setting loss weight for each loss component,
We employs a pragmatic multi-stage training methods, making the whole training process splitted into four stages.

In the first stage, only CFN is optimized. In detail, we fix parameters of all pretrained convolution layers in the body 
network (\layername{conv1\_1} to \layername{conv4\_3}), and drop all RPN and FRCNN layers to train the CFN.
In the second stage, we fix the whole body network (including the convolution layers for aggregating activation maps) 
and CFN, and train only RPN. 
Then in the third stage, body network, CFN and RPN are all fixed; only FRCNN component is optimized. 
While in the final stage, all layers are jointly optimized.

Acrossing all stages, in the optimization of the FRCNN, we treat region proposals coordinates from RPN as fixed value 
and do not back-propagate the gradient.

\section{Experiments and results}
The performance of HyperLearner is evaluated across multiple pedestrian datasets: KITTI~\cite{Geiger2012CVPR}, Caltech Pedestrian~\cite{dollar2009pedestrian}, and Cityscapes~\cite{Cordts2016Cityscapes}. The datasets we chose cover most of the popular ones in pedestrian detection task. 

One may also notice that our body network an implementation of HyperNet proposed in~\cite{kong2016hypernet}. 
Thus, we implement a control experiment where the CFN is removed as a typical HyperNet setting. 
That is, the body network keeps its side branches for aggregated activation map, but it does not learn from any extra
supervision.

\subsection{KITTI Dataset}
We evaluated the performance of HyperLearner with two kinds of feature supervision: edge and semantic segmentation.
These two kinds of channel features have been proved to be effective when directly integrated into the Faster R-CNN
framework (see Section~\ref{subsec:integration}). 
The results on the validation set of KITTI dataset is illustrated in the Table~\ref{tab:kittivalidation}.

\begin{table}[!tb]
\begin{center}
\setlength{\tabcolsep}{3pt}
\begin{tabular}{l || a c c | a c c }
\multirow{2}{*}{Model} & \multicolumn{3}{c|}{1x input} & \multicolumn{3}{c}{2x input} \\ \cline{2-7}
    & \multicolumn{1}{c}{Mod} & Easy & Hard & \multicolumn{1}{c}{Mod} & Easy & Hard \\ 
\hline
\hline
Fr-RCNN*~\cite{ren2015faster} & 59.29 & 64.53 & 53.01 & 71.05 & 76.00 & 62.08 \\
MS-CNN~\cite{cai2016unified} & 68.37 & 73.70 & 60.72 & 72.26 & 76.38 & \bfred{64.08} \\
\hline
Baseline    & 69.80 & 74.37 & 61.20 & 71.73 & 77.84 & 62.30 \\
HyperNet    & 69.72 & 76.91 & 61.10 & 72.23 & 77.96 & 63.43 \\
+Segmentation & 71.15 & 79.43 & \bfred{62.34} & 72.35 & \bfred{79.17} & 62.34 \\
+Edge       & \bfred{71.25} & \bfred{78.43} & 62.15 & \bfred{72.51} & 78.51 & 63.24
\end{tabular}
\end{center}
    \caption[]{Results on KITTI validation set, the model HyperNet refers to the HyperLearner without CFN.
    Evaluation follows moderate metric in KITTI. \\ *: Fr-RCNN follows setting as \cite{ren2015faster} while baseline model is Faster-RCNN with slightly different
    parameters. See also Table~\ref{tab:featurecomparison}.}
\label{tab:kittivalidation}
\end{table}

% In this figure, the columns with header .
In experiments on \emph{1x} scale, we notice great performance improvement when our HyperLearner is jointly learned from an 
edge detection network or a semantic segmentation network compared to the Faster R-CNN baseline and the 
HyperNet. The quantitative analysis is consistent with the experiments in Section~\ref{subsec:integration} 
where we directly integrate them as an extra input into the network through a branch network.

In experiments on \emph{2x} scale, HyperLearner as well as HyperNet make clear improvement. Based on former analysis, 
when the input image is of high resolution, the introduction of channel features with low-level details could benefit
the detector. 
% the introduction of channel features when the input image is of high resolution provides detailed low-level features to the network. 
In HyperNet setting, side branches of the body network act as an multi-level feature extractor,
and therefore such kind of improvement is expected.

As a transfer learning application, HyperLearner successfully boost a CNN-based detector using features learned by 
other networks with different architecture and trained for other tasks. 
% From another perspective, HyperNet-like CNNs has been proved to be effective in many vision tasks. 
From another perspective, HyperLearner 
offers an alternative way to perform feature learning in such CNNs and showed noticeable improvement.
Based on the results in Table~\ref{tab:kittivalidation}~and~\ref{tab:cityscapesvalidation}, it is safe to conclude that 
HyperLearner actually utilizes the extra supervision from channel 
features to generate a better hyper-feature extractor, especially for the detection task.

% Through former experiments, it is safe to conclude that HyperLearner actually utilizes the extra supervision from channel features to generate a better hyper-feature extractor, especially in the detection task.

\subsection{Cityscapes dataset}
The Cityscapes dataset~\cite{Cordts2016Cityscapes}, is a large-scale dataset for semantic urban segmentation which contains a diverse set of stereo video recordings from 50 cities. It consists of $2,975$ training and $500$ validation images with fine annotations, as well as another $20,000$ training images with coarse annotations. The experiments are conducted on the fine-annotated images. Compared with former standard datasets, Cityscapes possesses meticulous detection labeling (pixel-level), as well as fine semantic segmentation labeling.

As mentioned, the Cityscapes dataset provides pixel-level semantic segmentation labeling, so instead of using segmentation model pretrained on MS-COCO dataset, we directly address the multi-task learning by employing pixel-level segmentation labels as supervisor (\ie, our HyperLearner jointly learns pedestrian detection and semantic segmentation). 
During training, we only use segmentation labels for ``person''. 
As shown in Table~\ref{tab:cityscapesvalidation}, we also witness significant improvement over the Faster R-CNN baseline
and HyperNet.

\subsection{Caltech dataset}
The Caltech dataset~\cite{dollar2009pedestrian} is also a commonly used dataset for pedestrian detection evaluation. It consists of 2.5 hours 30Hz VGA video recorded from a vehicle traversing the streets of Los Angeles, USA. Detection results are evaluated on a test set consisting of 4024 frames.

\begin{figure}[!tb]
\begin{center}
   \includegraphics[width=0.9\linewidth]{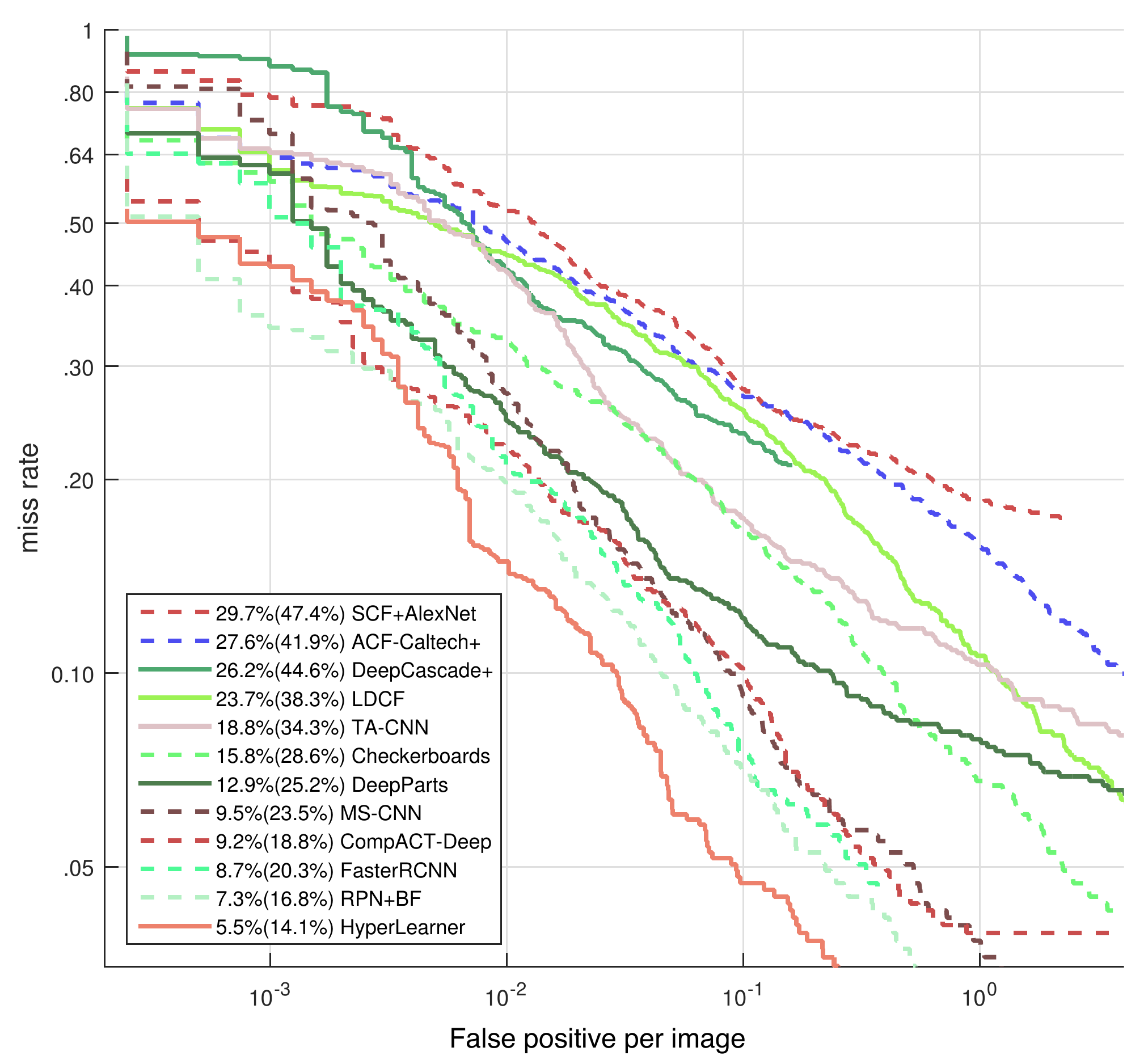}
\end{center}
   \caption{Detection quality on Caltech test set (reasonable, $\mathit{MR}_{-2}^N(\mathit{MR}_{-4}^N)$),
   evaluated on the new annotations~\cite{zhang2016far}. We achieve state-of-the-art 
   results on both evaluation metrics.
   }
\label{fig:caltech}
\vspace{-1em}
\end{figure}

Zhang \etal~\cite{zhang2016far} conducted a detailed survey and provided a refined groundtruth labeling on Caltech dataset. Our experiments is completely based on this new labeling (both training and testing). 
HyperLearner achieves state-of-the-art performance on the test set. Figure~\ref{fig:caltech} shows the detailed comparison of HyperLearner, the Faster R-CNN baseline and other methods.

\section{Summary}

In this paper, we integrated channel features into CNN-based pedestrian detectors, specifically, ICF channel, edge 
channel, segmentation channel and heatmap channel (apparent-to-semantic channel); optical flow channel (temporal channel); 
disparity channel (depth channel). 
Our quantitative experiments show semantic channel features can help detectors discriminate hard positive samples and 
negative samples at low resolution, while apparent channel features inhibit false positives of backgrounds and improve 
localization accuracy at high resolution. 

To address the issue of computational cost, we propose a novel framework, namely HyperLearner, to jointly learn channel 
features and pedestrian detection. HyperLearner is able to learn the representation of channel features while requiring 
no extra input in inference, and provides significant improvement on several datasets. From another point of
view, HyperLearner offers an alternative way to perform feature learning in HyperNet-like CNNs in a transfer
learning manner.

\begin{table}[!tb]
\begin{center}
\setlength{\tabcolsep}{3.5pt}
\begin{tabular}{l || c c | c c | c c}
\multirow{2}{*}{Model} & \multicolumn{2}{c|}{540p input} & \multicolumn{2}{c|}{720p input} 
                    &  \multicolumn{2}{c}{Improvement} \\ \cline{2-7}
& Speed & AP & Speed & AP & 540p & 720p \\ 
\hline
\hline
Baseline    & 130ms & 74.97 & 240ms & 86.89 & - & - \\
HyperNet    & 140ms & 74.30 & 250ms & 86.67 & -0.53 & -0.22 \\
Jointsegmap & 140ms & \bfred{77.22} & 250ms & \bfred{87.67} & \bfred{+2.25} & \bfred{+0.78}
\end{tabular}
\end{center}
    \caption{Results on Cityspcaes validation set. The speed column shows the time each model needed to 
    perform detection on a single image. The speed is tested on single NVIDIA TITAN-X GPU. We use all segmentation 
    polygons labeled ``person'' to generate bounding boxes for the pedestrian detection task. Following the standard 
    in Caltech dataset~\cite{dollar2009pedestrian}, all persons with (pixel-level) occlusion greater than 0.5 or of 
    height less than $50$ pixels are ignored. Furthermore, all polygons labeled ``cyclist'' or ``person group'' are 
    also ignored.
 }
\label{tab:cityscapesvalidation}
    \vspace{-0.5em}
\end{table}

\begin{figure}[!tb]
\begin{center}
   \includegraphics[width=0.95\linewidth]{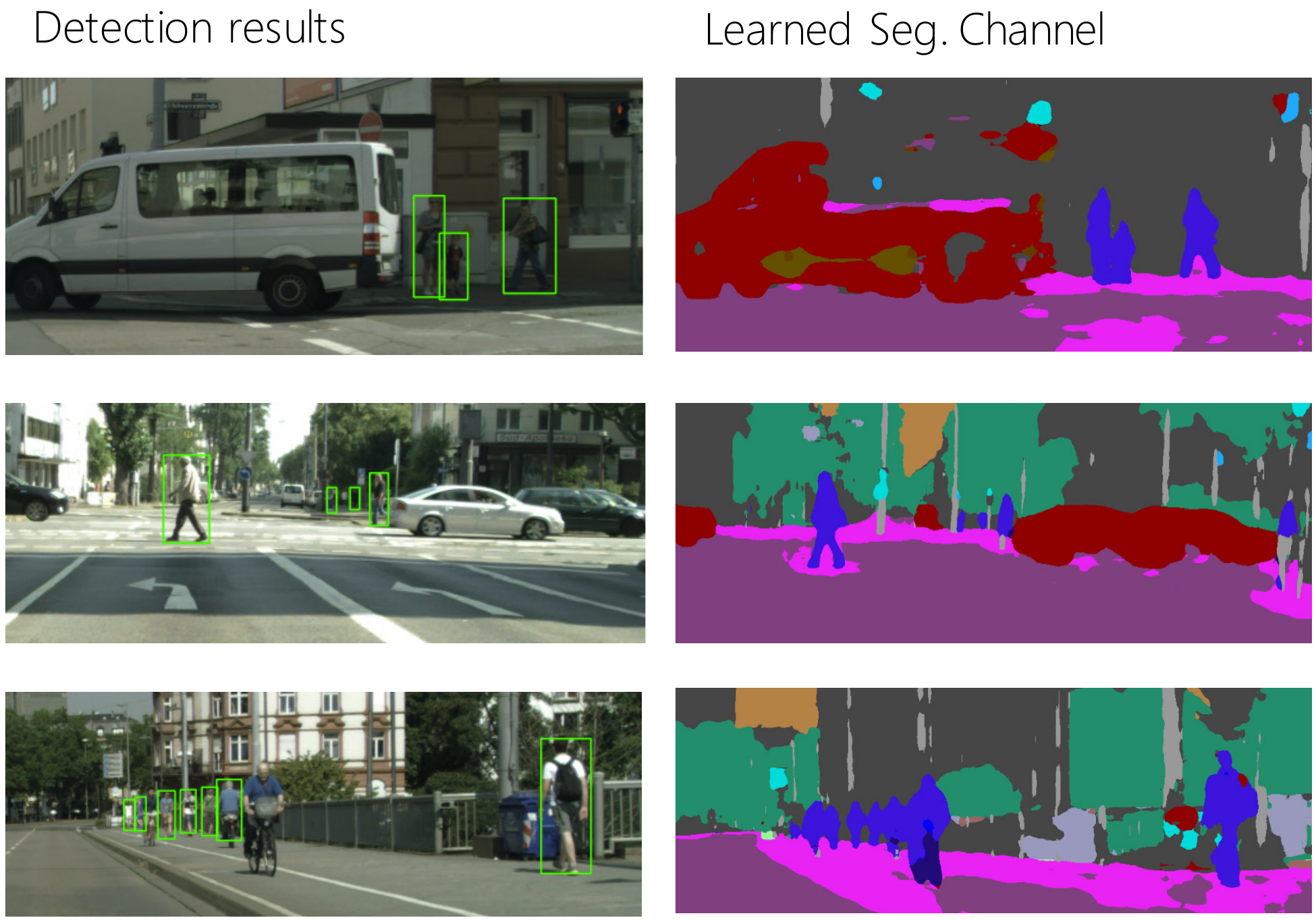}
\end{center}
   \caption{Results of HyperLearner on Cityscapes validation set. The left column shows our detection result, 
    while the right column demonstrate CFN's output learned from segmentation labeling.}
\label{fig:democityscapes}
    \vspace{-1em}
\end{figure}

\newpage
{\small
\bibliographystyle{ieee}
\bibliography{ref}
}

\end{document}